\documentclass[runningheads]{llncs}
\usepackage{xcolor}
\usepackage{algorithm}
\usepackage{algpseudocode}
\usepackage{bbold}
\usepackage{comment}
\usepackage{amsmath}
\usepackage{multirow}
\usepackage{amssymb}

\definecolor{darkgreen}{rgb}{0.0, 0.5, 0.0}
\DeclareMathOperator*{\argmin}{argmin}

\usepackage{graphicx}
\begin{document}
\title{Integrating Prior Knowledge in Post-hoc Explanations}
%
%
\author{Adulam Jeyasothy\inst{1}\orcidID{0000-0002-6375-5879} \and
Thibault Laugel\inst{2}\orcidID{0000-0002-5921-3225} \and
Marie-Jeanne Lesot\inst{1}\orcidID{0000-0002-3604-6647} \and  Christophe Marsala\inst{1}\orcidID{0000-0002-4022-9796} \and Marcin Detyniecki\inst{1,2,3}\orcidID{0000-0001-5669-4871 }}
\authorrunning{Jeyasothy et al.}
%
\institute{Sorbonne Université, CNRS, LIP6, F-75005 Paris, France \email{\{adulam.jeyasothy, marie-jeanne.lesot, christophe.marsala\}@lip6.fr} \and
AXA, Paris, France
\email{\{thibault.laugel, marcin.detyniecki\}@axa.com} \and
Polish Academy of Science, IBS PAN, Warsaw, Poland}
\maketitle              
\begin{abstract}
In the field of explainable artificial intelligence (XAI), post-hoc interpretability methods aim at explaining to a user the predictions of a trained decision model. Integrating prior knowledge into such interpretability methods aims at improving the explanation understandability and allowing for personalised explanations adapted to each user. 
In this paper, we propose to define a cost function that explicitly integrates prior knowledge into the interpretability objectives: we present a general framework for the optimization problem of post-hoc interpretability methods, and show that user knowledge can thus be integrated to any method by adding a compatibility term in the cost function. We instantiate the proposed formalization in the case of 
counterfactual explanations and propose a new interpretability method called Knowledge Integration in Counterfactual Explanation (KICE) to optimize it. The paper performs an experimental study on several benchmark data sets to characterize the counterfactual instances generated by KICE, as compared to reference methods.


\end{abstract}

\section{Introduction}
\label{sec : Introduction}

Machine learning models are now present in many areas of everyday life. The XAI domain~\cite{Burkart2021,Linardatos2021} aims at answering questions raised by these models such as "Why do I get this prediction?" or "What do I need to do to get the prediction I want?". In particular, post-hoc methods focus on generating explanations for the predictions obtained from a trained classifier. They come in different formats such as feature importance (e.g LIME~\cite{ribeiro2016} and SHAP~\cite{lundberg2017}) or counterfactual examples~\cite{wachter2018} (e.g. Growing Spheres~\cite{laugel2018} and FACE~\cite{Poyiadzi2020}). They also vary depending on their input data: model agnostic and data agnostic approaches consider that no knowledge is available either on the model, nor on data. This lack of knowledge makes it difficult to generate explanations adapted to the context. As a result, they may then not be always understood by the user~\cite{rudin2019}. To tackle this issue, some studies focus on enriching the considered input and integrating the human in the loop by taking into account expert knowledge~\cite{mahajan2019,frye2020,Ustun2019}, so the explanation is more understandable.

In this context, this paper proposes a general formalization, through the definition of a cost function that integrates such prior knowledge in the search for explanation. It proposes to add to the classical penalty term, that aims at assessing the quality of a candidate explanation, a complementary compatibility term that takes as parameter the considered knowledge. We provide a discussion about this notion of compatibility for which two contradictory semantics can be considered. Indeed, the prior knowledge can be used either to find an explanation that is complementary to the knowledge or an explanation that follows the same language and is in agreement with the knowledge. It can be argued that an explanation in the knowledge language can increase the user's confidence in the explanation, and a complementary explanation can enrich the user's knowledge. 

The paper discusses these cases when the considered knowledge provides information regarding features of interest. Then, it focuses on the case of explanation expressed within the knowledge language in the specific framework of counterfactual explanation: it considers that an explanation is compatible with the available knowledge if the features considered by the knowledge and the explanation are similar. In order to generate such a compatible explanation, it proposes a method called KICE which stands for Knowledge Integration in Counterfactual Explanation and presents the conducted illustrative experiments.

To summarize, the paper contribution consists in proposing a general framework for integrating expert knowledge into interpretability methods in~Section~\ref{sec:proposed_general_function}, an instantiation of the framework in the context of counterfactual explanations and a new counterfactual explanation method to generate a knowledge-adaptive explanation in Section~\ref{sec:proposition_CF}.

\section{Background}

This section provides background information regarding the framework to which the paper refers: it first describes some post-hoc explanation methods; then, it presents some existing approaches to integrate knowledge, after discussing different possible forms the latter can take.

\subsection{Post-hoc Explanations}
\label{sec: Post-hoc explanations}

Numerous post-hoc methods which explain the prediction of a trained classifier have been proposed, addressing different problems, see e.g.~\cite{guidotti2019}. A large portion of them relies on generating explanations as a solution to an optimization problem. Among those, we are particularly interested in counterfactuals and surrogates approaches. The former~\cite{lash2017,laugel2018,wachter2018,Poyiadzi2020,Mothilal2020} aim at answering the question: "What do I have to do to get the prediction I want?". The answer is defined as the closest instance to the studied instance belonging to the desired class. It can be obtained by optimizing, under constraint, a cost function defined as the distance function between the considered instance and the explanation (usually the Euclidean distance~\cite{lash2017,laugel2018}). On the other hand, surrogate-based explanations~\cite{ribeiro2016,peltola2018,sokol2019} rather answer the question: "Why do I get this prediction?". They rely on an interpretable model, such as a decision tree, a linear regression or classification rules, that approximates the classifier under study to mimic its behavior. In the case of local surrogates we can cite LIME~\cite{ribeiro2016} that minimizes the fidelity to the black box classifier. 

\subsection{Representation of Knowledge}

As discussed in the introduction, post-hoc explanations, as well as any explanation building method, can be enriched by available prior knowledge so as to improve their quality and intelligibility. The next section describes existing approaches, we discuss here several forms that can be considered to represent this knowledge; one can cite among others: class prototypes~\cite{VanLooveren2021}, distribution of data~\cite{Poyiadzi2020} or features-based knowledge~\cite{Ustun2019}. We are particularly interested in the latter, which is detailed in this section.

A first type of expert knowledge can be expressed by information on the individual features: the user can for instance indicate the so-called actionable features~\cite{Ustun2019}, i.e. the features that can be modified. For instance, they can include the budget in a credit application, as opposed to the age~\cite{lash2017}. A second type of expert knowledge on the features can provide information on their interactions through their covariation~\cite{drescher2013,mahajan2019} (e.g. "An increase in school level leads to an increase in age") or through a causality graph~\cite{mahajan2019,frye2020}. In this paper, we consider knowledge provided by a set of expected features that can for instance be deduced from a rule-based model. 

\subsection{Integration of Knowledge}
\label{sec : Integration_of_knowledge}

Beyond the question of the form the prior knowledge can take, the question of its exploitation and integration in the explanation must be considered. Existing works differ in the way they answer this question and the problem they address. 

Considering the case where knowledge indicates which features are actionable and explanations take the form of counterfactual examples, Ustun et al.~\cite{Ustun2019} restrict the search set to forbid the use of some features or directions. This allows to avoid proposing inapplicable explanation such as "In order to get the credit, you need to decrease your age". 

Also considering the case of counterfactual explanations, Mahajan et al.~\cite{mahajan2019} consider knowledge about causal interactions between features. They propose to integrate it in the definition of a new distance measure that quantifies the extent to which the candidate explanation satisfies the causal relationships. Unlike Ustun et al., they do not exclude candidates that do not respect these constraints, but penalize them, in a more flexible approach. This method makes it possible to avoid unrealistic explanations of the form "In order to get the credit, you need to increase your budget and to decrease your salary".

Frye et al.~\cite{frye2020} are also interested in the integration of causal relationships, but exploit a different information they induce; they considers the case of explanations in the form of local feature importance vector. He focuses on the asymmetry brought by the causal link orientation and proposes to exploit it by weighting the cost function. 

These three approaches integrate different types of expert knowledge into different forms of explanation. They consider that this knowledge is an additional constraint in the explanation generation. This constraint is represented in different forms: reduction of the search space, addition of penalty in the cost function or weight in the cost function. In the following section, we propose a general formalization for all methods integrating knowledge. 

\section{Proposed General Framework}
\label{sec:proposed_general_function}

The considered objective is to explain the prediction of a classifier $f: \mathcal{X} \longrightarrow \mathcal{Y}$ where $\mathcal{X}$ denotes the input space, included in $\mathbb{R}^d$, and $\mathcal{Y}$ the output space. We consider  a domain where elements $x \in \mathcal{X}$ are described by means of a set of features (for instance, age, salary, weight,...). In this paper, we consider continuous data, the case of categorical data will be considered in future work. The prior knowledge to be integrated in the generation of the explanation is denoted~$E$. In the following, we consider that an explanation is related to the prediction $f(x)$ for an element $x \in \mathcal{X}$.

\subsection{Proposed Optimization Problem}
This section describes the proposed general optimization problem to generate explanation integrating user knowledge. 
Let $\mathcal{E}$ be the set of explanations for a given type~(e.g. counterfactual example or surrogate models), we propose the following optimization problem: 
\begin{equation}
\argmin_{e\in \mathcal{E}} agg(penalty_x(e),incompatibility_x(e,E))
\label{eq:framework_general} 
\end{equation}
where $agg$, $penalty_x$ and $incompatibility_x$ are three functions described in the following.
This optimization problem allows to generate a knowledge-based explanation for any type of explanation and type of knowledge. As discussed below, the choice of the three functions depends on the studied context: the motivations of the user, the type of explanation to be generated or the considered type of knowledge. In our approach, we use a subscript $x$ for functions that are related to the studied instance $x$. 

\subsubsection{Penalty Function}

As recalled in Section~\ref{sec: Post-hoc explanations}, most existing approaches to generate explanations minimize a cost function. Let $penalty_x : \mathcal{E} \longrightarrow \mathbb{R}$ be such a function: it takes as argument a candidate explanation $e$, and may depend on the studied instance~$x$ in the case of local explainer. It values the quality of the candidate explanation. For instance, it is defined as the distance between the candidate and the considered instance in the case of counterfactual examples and the fidelity of explanation to the black box classifier in the case of surrogate approaches.

\subsubsection{Incompatibility Function}
\label{sec : Proposition_incompatibility}

To generate an explanation complying with available knowledge, we propose to add a function to measure how compatible the candidate explanation is with the knowledge. This incompatibility function depends on the type of explanation and the type of considered knowledge, as well as the user objective.
For instance, an explanation generated for a non-expert customer asking how to get his credit application accepted needs to focus on common knowledge actions such as increasing his salary. On the other hand, in the case where the explanation is destined to a domain expert, the incompatibility can represent the level of fidelity of the explanation with respect to the expert's comprehension of the domain.

As mentioned in the introduction, we propose to distinguish two objectives: to propose an explanation that is complementary to the available knowledge, and to propose an explanation that compete with the knowledge language. We consider an explanation to be in the knowledge language if the features used in the explanation and the knowledge are similar, and an explanation to be complementary in the opposite case (i.e. when there is no redundancy). Below, we discuss these two possibilities 
in the case of surrogate-based explanations. 

Let $A_e$ denote the set of features used in the explanation generated by a surrogate $e$ which is an interpretable model, and $\overline{E}$ the set of features not considered in knowledge $E$. To build a surrogate-based explanation in the knowledge language, one possibility is to minimize the number of features it takes into account that are not part of~$E$:
\[ incompatibility_{x}(e,E) = Card(A_{e} \cap \overline{E}) \]
On the contrary, where the explanation needs to be complementary to the knowledge, another possibility is to minimize the number of features present in both~$A_{e}$ and~$E$:
\[ incompatibility_{x}(e,E) = Card(A_{e} \cap E) \]
These two definitions can be both relevant depending on the motivations and needs of the user, as discussed above. Such a case is further discussed in Section~\ref{sec:proposition_CF} for the case of counterfactual explanation.

\subsubsection{Aggregation Function}

The aggregation function combines the penalty and incompatibility functions. There exists an abundant literature on aggregation operators, see e.g.~\cite{Mesiar02,Grabisch09}, which can be divided into three main categories: conjunctive, disjunctive or compromise. A conjunctive operator, for example the min function, returns a high value only if both penalty \emph{and} incompatibility taken as input are high. A disjunctive operator can be represented by the max function, it requires \emph{at least one} value to be high. Finally, trade-off functions such as weighted average result in the compensation of low values by high values. 

The choice of this function can be made according to the user preferences, instead of necessarily considering the same function for everyone: one way to personalize the explanation to a user is to let him choose the aggregation he wants to perform. 

\subsection{An Example of Existing Method under this Formalism}

To show how general the proposed framework is, we propose to express the state-of-the-art approach proposed by Ustun et al.~\cite{Ustun2019} in the proposed formalism 
highlighting the definition of the three involved functions. 
In this approach, the user knowledge~$E$ is denoted $A(x)$ and defined as the set of modifications that can be applied to instance $x$, it is integrated to generate an actionable counterfactual explanation. The latter is the solution to an optimization problem of the form: 
\[ a^* = \argmin_{a \in A(x)} cost\_fct(a,x) \text{ with }f(x+a) \neq f(x) \]
where $cost\_fct$ values the distance between $x$ and $x+a$. The considered explanation expresses the move between the studied instance and the closest instance in the opposite class. This optimization problem can be rewritten as follows: let $\mathcal{E}_x = \{x' \in \mathbb{R}^{d}\; |\; f(x')~\neq~f(x)\}$,
\[e^* = \argmin_{e \in \mathcal{E}_x} cost\_fct(e,x) + \mathbb{1}_{|x-e| \notin A(x)} \times Z \]
where $Z$ is arbitrarily large. Here $e^*$ is the closest instance in the opposite class, and it enables to retrieve $a$ as $a=e^*-x$. 

This expression, equivalent to the previous one, makes it possible to identify the $penalty$, $incompatibility$ and $agg$ functions. The penalty function equals $cost\_fct$ defined in~\cite{Ustun2019}. The incompatibility function equals $\mathbb{1}_{|x-e| \notin A(x)} \times Z $ and represents  the presence or absence of the modified feature in the user knowledge; it takes only two values 0 or Z. An incompatible counterfactual explanation thus has a very high cost function, resulting in only compatible counterfactuals being considered. Finally, aggregation is performed by a weighted sum. However, as the incompatibility function is binary, only compatible counterfactuals are considered, so the resulting counterfactual is both compatible and of good quality. 

\section{Knowledge Integration in Counterfactual Explanation (KICE)}
\label{sec:proposition_CF}

This section proposes a new method to generate counterfactual explanations taking into account prior knowledge, instantiating for counterfactuals the general framework introduced in the previous section. It then details the algorithm used to solve this problem. In the following, to value a distance we consider the $l_2$ norm.

The principle of the expected counterfactual example can be illustrated in the following toy scenario; let us consider the case of a patient who risks diabetes if he keeps his daily habits. Without knowledge integration, a plausible explanation is: "To predict a low risk, the patient has to decrease his blood sugar by 0.5g/L and increase his carbohydrate by 50g". However, the knowledge that the patient possesses is only reduced to less technical features, such as  \textit{sport duration} and \textit{weight}: the proposed counterfactual explanation is thus not understandable to him. We aim to propose a counterfactual explanation more adapted to the user knowledge: "The patient has to run 30 minutes longer and lose 2kg". 

\subsection{Proposed Cost Function: Instantiation of the Framework}

We propose in this section an instantiation of the general framework in the counterfactual case and considering prior knowledge in the form of a set of features denoted $E = \{a_i, i= 1, ..., m\}$, considered as a subset of features of the input space~$\mathcal{X}$.

\label{sec : Proposed_cost_function}

\subsubsection{Penalty Function} 
For the penalty function, we propose to use the classical counterfactual function which is the Euclidean square distance: 
\begin{equation}
    penalty_{x}(e) = \parallel x - e \parallel^2
    \label{eq:penalty}
\end{equation}  

\subsubsection{Incompatibility Function} As specified in Section~\ref{sec : Proposition_incompatibility}, the objective is to propose a counterfactual explanation in agreement with the user knowledge, meaning that ideally, the counterfactual modifications must be only performed according to the features appearing in $E$. 
However, for some instances, focusing only on a subset features raises the risk of never being able to meet the decision boundary of $f$, leading to no counterfactual explanation being generated. Therefore, we propose to relax this constraint by penalizing the modifications according to the features from $\overline{E}$ that are not present in the knowledge. This allows, when the  boundary cannot be found along the sole features of $E$, to make sure that a solution can still be found. Thus, we propose the incompatibility function that computes the Euclidean square distance only along absent features:
\begin{equation}
incompatibility_{x}(e,E) = \parallel x - e \parallel_{\overline{E}}^2 = \sum_{i \notin E} (x_i - e_i)^2
\label{eq:incompatibility}
\end{equation} 
Minimizing this incompatibility avoids generating counterfactual explanations that greatly modify the unknown features. 

\subsubsection{Aggregation Function} The aggregation function combines the penalty function and the incompatibility function, we propose to use a compromise function:
\begin{equation}
    agg(u,v) =  u + \lambda v
\end{equation}
with $\lambda \in \mathbb{R}^{+}$ a hyper-parameter valued by the user. $\lambda$ defines the weight given to the incompatibility function. In the case of categorical features, the $l_2$ distance is replaced with an appropriate distance or a dissimilarity measure.

We then obtain the following optimization problem: 

Given  $\mathcal{E} = \{x' \in \mathbb{R}^{d} \; | \; f(x')~\neq~f(x)\}$ and $\lambda \in \mathbb{R}^+$,

\begin{eqnarray}
\label{eq:fct_opti}
e^* &=& \argmin_{e \in \mathcal{E}} cost_{x,E}(e) \\
\text{with : } 
\label{eq:fct_cost}
cost_{x,E}(e) &=& \parallel x - e \parallel^2 + \lambda \parallel x - e \parallel_{\overline{E}}^2
\end{eqnarray}

\subsection{Algorithm Description} 

The algorithm we propose to solve optimization problem~(\ref{eq:fct_opti}), named KICE for Knowledge Integration in Counterfactual Explanation, uses the principle of iterative generation of instances, inspired by the Growing Spheres algorithm~\cite{laugel2018}, considering the additional term of compatibility to bias the generation. We consider a model and data agnostic case, in which no information about the data distribution nor the decision boundary is available. Therefore, we generate instances around the considered instance~$x$ (i.e. for which explanation related to~$f(x)$ should be found) in all directions: to find the point that minimizes the cost function, we generate points in increasingly larger  spaces until we find one in the opposite class. 

The equation $cost_{x,E}(e) = \nu$ defines the equation of an ellipse with center~$x$ and radius~$\sqrt{\frac{\nu}{1+\lambda}}$ with respect to the features in~$\overline{E}$ and $\sqrt{\nu}$ with respect to the features in~$E$. To make the ellipses grow, the value taken by $\nu$ is iteratively increased. As we cannot cover all possible values for $\nu$, a hyper-parameter $\epsilon>0$ is used to generate elliptical layers defined by radius $\nu$ and $\nu + \epsilon$ and generate uniformly in these layers. 

In order to do so, we use a modified version of the HLG algorithm~\cite{mueller1959} which consists in generating instances uniformly in the spherical layer $SL(x, a_0, a_1)$ which represents the set of points at a distance greater than~$a_0$ and less than~$a_1$ from~$x$. In the HLG algorithm the rescaling step consists in generating a random variable $U~=~\{u_i\} \sim \mathcal{U}([0,1])$, and then calculating $a_0 + a_1U^{\frac{1}{dim(\mathcal{X})}}$. We distinguish the features $E$ and $\overline{E}$, and perform the following operations: $a_0 + a_1u_i$ for~$i \in E$ and $\frac{a_0}{\sqrt{1+\lambda}} + \frac{a_1}{\sqrt{1+\lambda}}u_i$ for~$i \in \overline{E}$. The instances are then indeed generated in the desired ellipse.

\begin{figure}[t]
    \centering
    \includegraphics[scale=0.56]{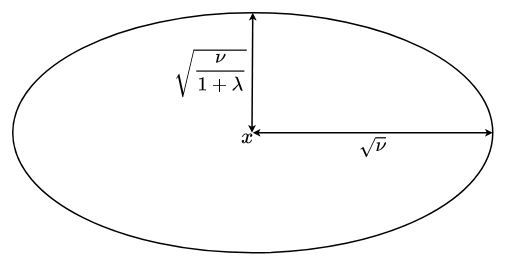}
    \includegraphics[scale=0.36]{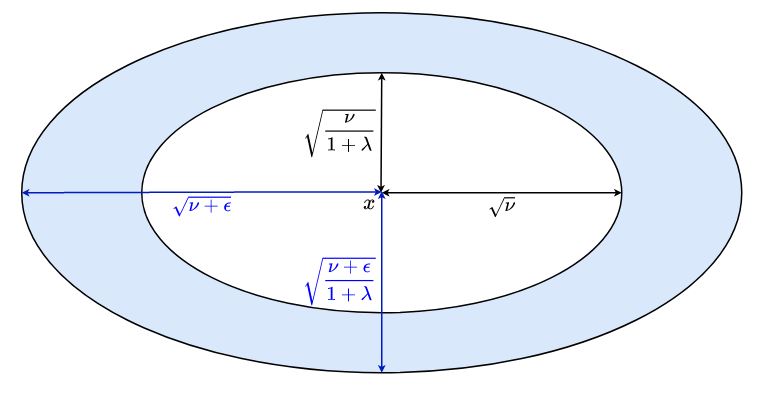}
    \caption{(left) Generation zone for the initial step, (right) Iteration zone.}
    \label{fig:ellipse}
\end{figure}

The proposed KICE algorithm covers the space by generating instances iteratively: in the first step, $n$ instances are generated in the ellipse of center~$x$ and radius $\sqrt{\frac{\nu}{1+\lambda}}$ with respect to the features in $\overline{E}$ and $\sqrt{\nu}$ with respect to the features in $E$. This ellipse is represented in the left part of Figure~\ref{fig:ellipse}. If none of these instances belong to the opposite class, we generate instances in the ellipsoidal layer between $\nu$ and $\nu+\epsilon$. This layer is represented in blue on Figure~\ref{fig:ellipse}.

\section{Experiments}

This section presents the experiments conducted to evaluate the KICE algorithm: the purpose of these experiments is to show that the proposed method finds the expected counterfactual instance, thus it allows to achieve a trade-off between the quality and the compatibility of an explanation. 

\subsection{Experimental Protocol}

Experiments are conducted on three classical benchmark tabular datasets: half-moons, Boston and breast cancer. As a pre-treatment, dataset values are normalized and in the case of the Boston dataset, the regression value is transformed into a binary class by means of a discretization step: the price is "expensive" if it is greater than $\$21,000$, and "cheap" otherwise. Datasets are split into train and test data subsets (80\%-20\%). Within the considered post-hoc explanation framework, the classifier choice does not matter, we apply SVM with a Gaussian kernel, that obtains good accuracy on the three datasets: half-moons (0.99), Boston (0.98) and breast cancer (0.93).  
Regarding prior knowledge, we consider the user disposes of less features than the classifier. In order to build somehow realistic knowledge, we train a decision tree with low depth on the train data (more precisely maximal depth is set to half the number of data features), the set of features $E$ is then the ones that occur in this tree. Counterfactual instances~$e^*$ are then generated for each instance $x$ of the test data set using KICE, with the value of $\lambda$ being chosen to observe interesting results, respectively $4$, $1$ and $6$ for the datasets half-moons, Boston and breast cancer.

We compare the proposed method KICE to two competitors. The first one is Growing Spheres~\cite{laugel2018}, which solves the reference counterfactual optimization problem that only minimizes the Euclidean distance and denote $e_{ref}$ the generated point. This corresponds to an extreme case of aggregation of Eq.~\ref{eq:fct_opti} where the incompatibility term is ignored i.e. $\lambda=0$. 
A second competitor is proposed by imposing to strictly comply with the knowledge. Its associated cost function thus minimizes the Euclidean distance according to the knowledge features only, a naive way of integrating knowledge in the explanation. 

We denote $e_{user}$ the counterfactual instance that solves the associated problem: 
\[e_{user} = \argmin_{e \in \mathcal{E}} \parallel x-e \parallel_{E}^2  \]
We can write this optimization problem in the proposed formulation given in Eq.~(\ref{eq:fct_cost}) with $\lambda$ arbitrarily large:
\[ e_{user} = \argmin_{e \in \mathcal{E}} \parallel x-e \parallel^2 + \lambda \parallel x-e \parallel_{\overline{E}}^2 \]

\subsection{Illustrative Examples}

\begin{figure}[t]
    \centering
    \includegraphics[width=0.32\textwidth]{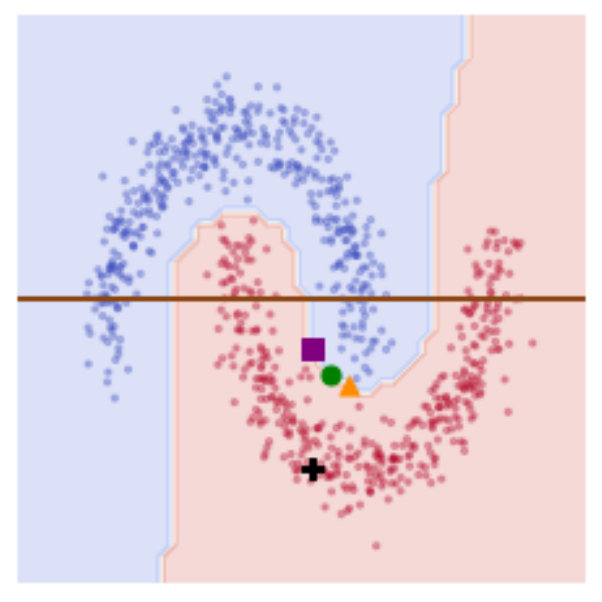}
    \includegraphics[width=0.32\textwidth]{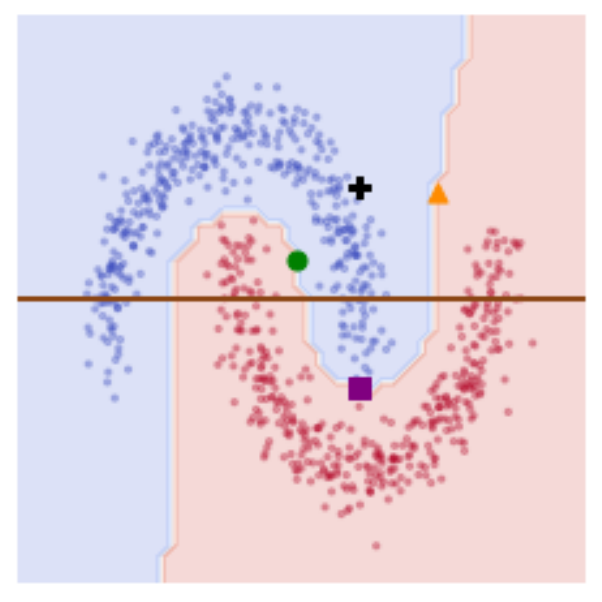}
    \includegraphics[width=0.32\textwidth]{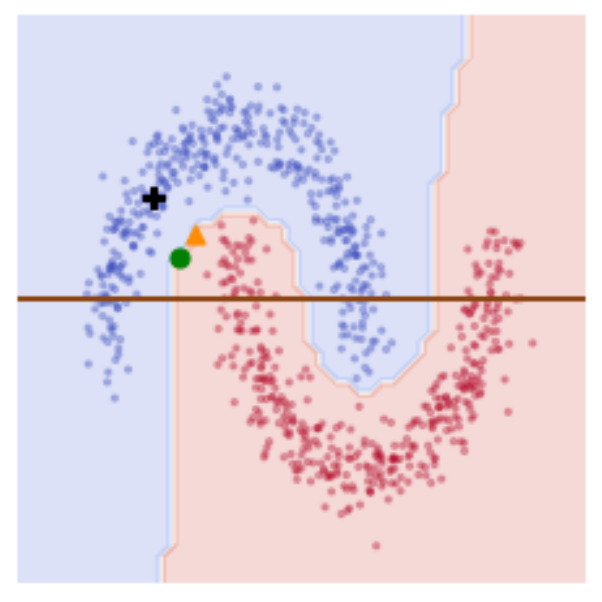}
    \caption{Examples of obtained results $e_{ref}$, $e_{user}$ and $e^*$ for three instances $x$ ($\pmb{+}$:~$x$, \textcolor{orange}{$\blacktriangle$}:~$e_{ref}$, \textcolor{violet}{$\blacksquare$}:~$e_{user}$, \textcolor{darkgreen}{$\bullet$}:~$e^*$)}
    \label{fig:half_moons_examples}
\end{figure}

First, we illustrate the behavior of the methods with examples in two dimensions denoted $X_0$ and $X_1$. Figure~\ref{fig:half_moons_examples} shows the counterfactual examples obtained for three different instances of the half-moons dataset. The figure shows the training set, the blue and red regions represent the predicted classes (darker points are the training examples); the decision boundary of the trained SVM classifier is shown in white, it achieves 0.99 accuracy. The considered knowledge system is a rule on a single feature, it is represented by the brown horizontal line : $E = \{X_1\}$. 

In the figure, we observe that the counterfactual instance~$e_{ref}$ is the closest point belonging to the opposite class. As expected, $e_{user}$ is further away than $e_{ref}$ and only modifies feature~$X_1$ which is the knowledge feature: we observe that $e^*$ is a compromise between~$e_{ref}$ and $e_{user}$. It requires less modifications according to $X_0$ than $e_{ref}$, hence it is more compatible. It is also closer to the studied instance than $e_{user}$. In the graph, on the right, we notice that there is no $e_{user}$. In this case, there is no counterfactual instance modifying only feature~$X_1$. The proposed method is then useful because it allows to obtain a more compatible counterfactual than~$e_{ref}$.

\subsection{Evaluation of the KICE Method}

In this section, we further compare the proposed method with its presented competitors and study quantitatively the gain in terms of the proposed cost function offered by KICE. 

We apply the three methods described in the previous section on the test set of the different datasets. Among the test set, some instances do not find counterfactual examples with the method that modifies only the features in~$E$ ($e_{user}$). 
For the datasets half-moons, Boston and breast cancer, this concerns respectively 20\%, 0\% and 11\% of the instances.
For the rest of the instances, Table~\ref{tab:Results} shows the mean and standard deviation values for the penalty, incompatibility and cost functions for the three approaches. 

\begin{table}[t]
    \centering
    \begin{tabular}{|c|c||c|c|c|}
        \cline{3-5}
         \multicolumn{2}{c|}{} & $penalty_{x}(e)$ & $incompatibility_x(e,E)$ & $cost_{x,E}(e)$ \\
        \cline{3-5}
        \hline
        \multirow{3}{*}{Half-moons}& $e_{ref}$ & \textbf{0.32 $\pm$ 0.21}& 0.14 $\pm$ 0.13 & 0.86 $\pm$ 0.56\\
        \cline{2-5}
        & $e_{user}$ & 1.48 $\pm$ 1.3 & \textbf{0.0 $\pm$ 0.0}& 1.48 $\pm$ 1.3\\
        \cline{2-5}
        & $e^*$ & 0.42 $\pm$ 0.29 & 0.08 $\pm$ 0.11 & \textbf{0.73 $\pm$ 0.52}\\
        \hline
        \hline
        \multirow{3}{*}{Boston}& $e_{ref}$ & \textbf{1.48 $\pm$ 1.75} & 0.7 $\pm$ 1.03 & 3.57 $\pm$ 4.72 \\
        \cline{2-5} & $e_{user}$ & 2.26 $\pm$ 2.71 & \textbf{0.0 $\pm$ 0.0} & 2.26 $\pm$ 2.71 \\
        \cline{2-5} & $e^*$ & 1.72 $\pm$ 2.09 & 0.13 $\pm$ 0.19 & \textbf{2.12 $\pm$ 2.54} \\
        \hline
        \hline
        \multirow{3}{*}{Breast cancer}& $e_{ref}$ & \textbf{8.82 $\pm$ 9.22} & 7.27 $\pm$ 8.25 & 52.41 $\pm$ 58.63\\
        \cline{2-5}
        & $e_{user}$ & 22.42 $\pm$ 24.87 & \textbf{0.0 $\pm$ 0.0} & 22.42 $\pm$ 24.87\\
        \cline{2-5}
        & $e^*$ & 10.74 $\pm$ 9.85 & 1.25 $\pm$ 1.33 & \textbf{18.24 $\pm$ 16.83} \\
        \hline
    \end{tabular}
    \caption{Results with the three considered approaches on the three datasets for metrics: $penalty$ defined in Equation~(\ref{eq:penalty}), $incompatibility$ defined in Equation~(\ref{eq:incompatibility}) and $cost$ defined in Equation~(\ref{eq:fct_cost})}
    \label{tab:Results}
\end{table}

We notice that, as expected, the proposed counterfactual examples have a penalty greater than that of $e_{ref}$ but lower than that of $e_{user}$. Moreover, the incompatibility function is much smaller than that of $e_{ref}$. Finally, $e^*$ cost function is the lowest. We notice that the standard deviations are high, which is due the fact that the instances are at different distances from the boundary. 

\begin{figure}[t]
    \centering
    \includegraphics[width=0.45\textwidth]{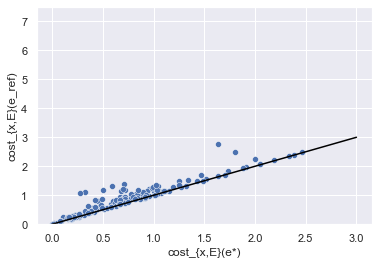}
    \includegraphics[width=0.45\textwidth]{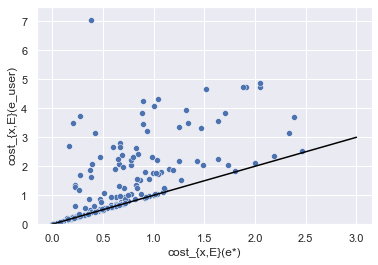}
    \caption{Cost functions $cost$ as defined in Equation~(\ref{eq:fct_cost}) of $e_{ref}$, $e_{user}$ and $e^*$ for the 80\% of the test set for which  all three counterfactual instances are defined.}
    \label{fig:cost_function}
\end{figure}

To verify that KICE  minimizes the cost function as opposed to the other two  on all instances, Figure~\ref{fig:cost_function} shows the value of the cost function obtained by $e^*$ (that is defined as minimizing it) as compared to the value it takes for $e_{ref}$ (left) and $e_{user}$ (right), for each of the test instances of the half-moons dataset. The figures show that, as expected, all points are above the line $y=x$.

On the right hand graph, the points are more scattered but they remain above the line. We notice that the generated counterfactual instances are closer to the cost function of $e_{ref}$ than to $e_{user}$. Only one feature is consider by the knowledge, it is difficult to get a close counterfactual instance only by modifying this feature. It is then necessary to move away from the studied instance to get 100\% compatible one. It is also possible to increase $\lambda$ so that the points are less scattered. 

\section{Conclusion and Perspectives}
\label{sec : Conclusion}

In this paper, a general framework is proposed to help defining an optimization problem to integrate prior knowledge in post-hoc model-agnostic explanations. Using this framework, we proposed a new method, KICE, to generate counterfactual explanations in the knowledge language by minimizing the modifications according to the unknown features. 

Future works will include a study of the counterfactuals explanations generated for different values of $\lambda$. Thus, we will study the $\lambda$ associated with the expected counterfactual examples. Another direction will focus on exploring and proposing instantiations of the framework for different models and applications, as well as real world experiments including real users. 

\bibliographystyle{splncs04}

\end{document}